\documentclass[10pt,twocolumn,letterpaper]{article}
\usepackage{cvpr}
\usepackage{times}
\usepackage{epsfig}
\usepackage{graphicx}
\usepackage{amsmath}
\usepackage{amssymb}
\usepackage{caption}
\usepackage{subfigure}
\usepackage{multirow}
\usepackage{cite}
\DeclareMathOperator{\atantwo}{atan2}

 \cvprfinalcopy 

\def\cvprPaperID{2026} 

\ifcvprfinal\fi
\begin{document}

\title{DDNet: Cartesian-polar Dual-domain Network for the Joint Optic Disc and Cup Segmentation}

\author{Qing Liu\\
Central South University\\
{\tt\small qing.liu.411@gmail.com}
\and
Xiaopeng Hong\\
University of Oulu\\
{\tt\small Xiaopeng.Hong@oulu.fi}
\and
Wei Ke\\
Carnegie Mellon University\\
{\tt\small weik@andrew.cmu.edu}
\and
Zailiang Chen\\
Central South University\\
{\tt\small xxxyczl@csu.edu.cn}
\and
Beiji Zou\\
Central South University\\
{\tt\small bjzou@csu.edu.cn}
}
\maketitle

\begin{abstract}
   Existing joint optic disc and cup segmentation approaches are developed either in Cartesian or polar coordinate system. However, due to the subtle optic cup, the contextual information exploited from the single domain even by the prevailing CNNs is still insufficient. In this paper, we propose a novel segmentation approach, named Cartesian-polar dual-domain network (DDNet), which for the first time considers the complementary of the Cartesian domain and the polar domain. We propose a two-branch of domain feature encoder and learn translation equivariant representations on rectilinear grid from Cartesian domain and rotation equivariant representations on polar grid from polar domain parallelly. To fuse the features on two different grids, we propose a dual-domain fusion module. This module builds the correspondence between two grids by the differentiable polar transform layer and learns the feature importance across two domains in element-wise to enhance the expressive capability. Finally, the decoder aggregates the fused features from low-level to high-level and makes dense predictions. We validate the state-of-the-art segmentation performances of our DDNet on the public dataset ORIGA. According to the segmentation masks, we estimate the commonly used clinical measure for glaucoma, i.e., the vertical cup-to-disc ratio. The low cup-to-disc ratio estimation error demonstrates the potential application in glaucoma screening.
   
   
\end{abstract}
\section{Introduction}
Automated segmentation of the optic disc (OD) and optic cup (OC) in the retinal fundus images is a fundamental task in the field of medical image analysis. It helps the quantification of the clinical measures about the retinal related diseases, such as the rim thickness, the ISNT rule \cite{ISNT2006}, and the vertical cup-to-disc ratio (CDR) \cite{CDR_IOVS_2000}. These measures further assist in the diseases diagnosis and the progression assessment, and facilitate for the doctor-patient communication. 
\\ \indent In the fundus, the OD consists of two parts: the OC exhibiting as a pit in centre and the neuroretinal rim packing the nerve fibres. Thus, a reliable feature to segment the OC and the rim is the depth. However, in $2$D images, the depth information is completely absent. This makes the OC segmentation problem be highly ill-defined.
\\ \indent The current consensus on the segmentation problem is to learn good representations for the OC pixels, rim pixels, and the background pixels. The deep features are now textbook. For example, U-shaped networks are designed in  \cite{Sevastopolsky_PRIA_2017} and  \cite{DenseNet_Systems_2018} to learn deep features in Cartesian domain. An MNet \cite{Fu_TMI_2018} is designed to learn deep features in polar domain. However, the representations learned from single domain are still insufficient to distinguish the OC pixels, rim pixels, and background pixels.
\begin{figure*}
\begin{center}
\includegraphics[width=\textwidth]{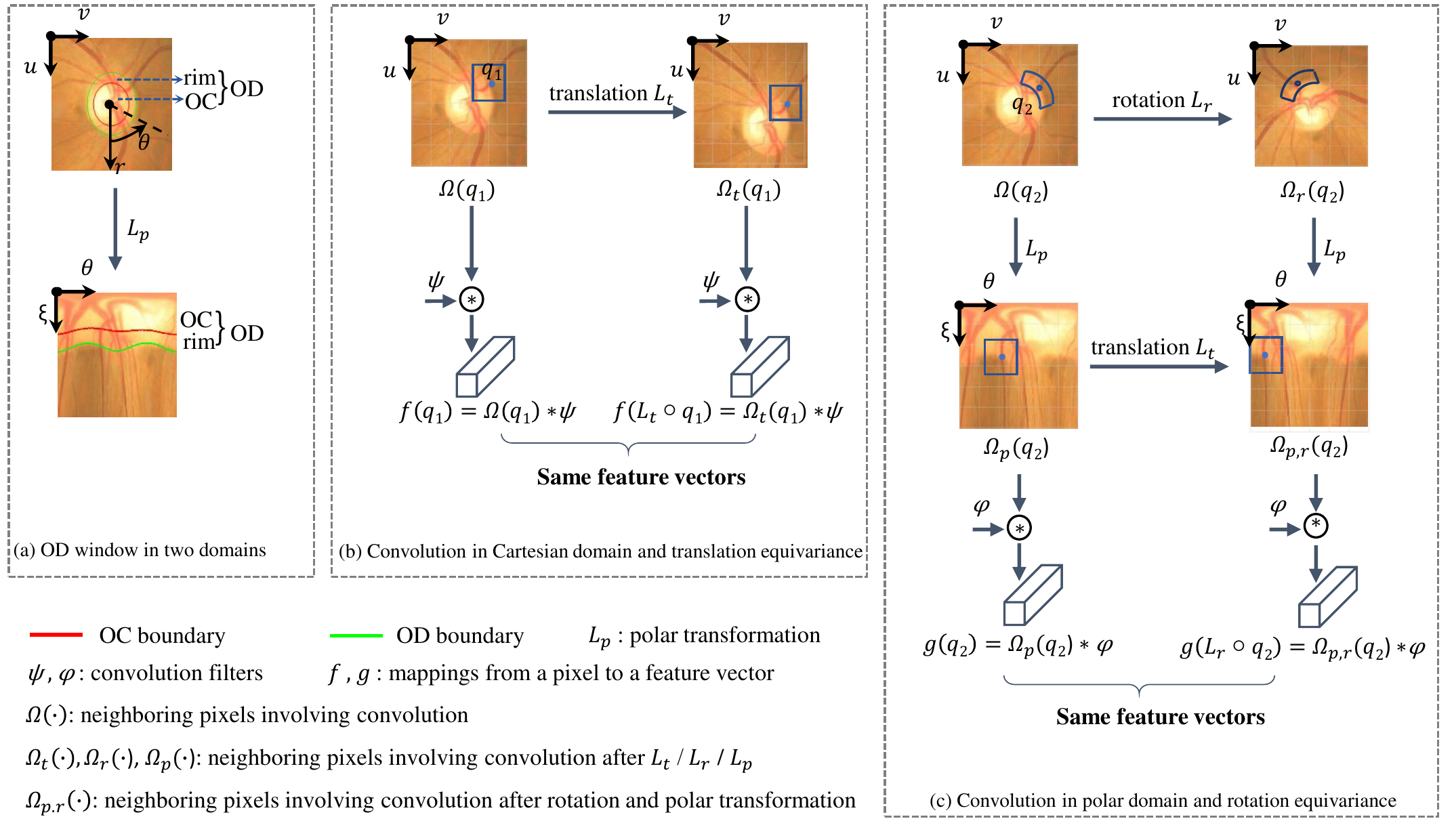}
   \caption{Illustration for the complementary of Cartesian domain and polar domain. (a) OD window in Cartesian domain (top) and polar domain (bottom). The differences on the shapes and spatial layout of the structures imply that two domains embed different contextual information. $(u,v)$ and $(\xi,\theta)$ denote the Cartesian coordinates and polar coordinates respectively. (b) and (c): Convolution in Cartesian domain and polar domain respectively. In Cartesian domain, the convolution is translation equivariance but not rotation equivariance. As illustrated in (b), for any pixel $q_1$ in the original image and its corresponding pixel denoted by $L_t\circ q_1$ in the image with translation $L_t$, convolution with a bank of arbitrary filters $\psi$ produces the same feature vectors.  Convolution in polar domain is rotation equivariance but not translation equivariance. As illustrated in (c), the rotation $L_r$ in Cartesian domain is reduced to translation $L_t$ in polar domain. Thus for any pixel $q_2$ in the original image and its corresponding pixel denoted by $L_r\circ q_2$ in the image with rotation $L_r$, transforming them into polar domain, then convolution with a bank of arbitrary filters $\phi$ produces the same feature vectors.}
\label{fig:exam4two_sys}
\end{center}
\end{figure*}
\\ \indent In this paper, we argue that integrating representations by CNNs from both the Cartesian and polar domains contributes to the accurate segmentation of the OD and OC. Intuitively, the shapes and spatial layouts of the OC, rim, and vessels are completely different in these two domains, as shown in Fig. \ref{fig:exam4two_sys} (a). This implies that different contextual information will be learned from images in different domains. Naturally, we are motivated to achieve richer representations for the segmentation by exploiting complementary contextual information from both domains and integrating them.
\\ \indent Theoretically, the CNNs in Cartesian domain are equivariant to translation \cite{cohen2016group}. More specifically, for any pixel in an image, the feature vector learned by the CNNs in Cartesian domain is translation invariant, as illustrated in Fig. \ref{fig:exam4two_sys}(b). On the other hand, in polar domain the CNNs are equivariant to rotation \cite{PTN_ICLR}. When considering one pixel of an image, the feature vector learned by the CNNs in polar domain is rotation invariant, as illustrated in Fig. \ref{fig:exam4two_sys}(c). Fusing translation equivariant representations from Cartesian domain and the rotation equivariant representations from the polar domain reaches a richer representation with high expressive capability and better predictive performances than any one of them.
\\ \indent In this paper, we propose a Cartesian-polar dual-domain network (DDNet) for the joint OD and OC segmentation. It first learns feature representations from both Cartesian domain and polar domain by a two-branch of domain encoder. Then it fuses the representations from two domains by the proposed dual-domain fusion module. The fusion module builds the correspondence between rectilinear grid and polar gird by the differentiable polar transform layer and learns complementary contextual information from domain feature maps. Finally, the fused features are used for the dense classification by a decoder. 
\\ \indent In summary, there are three contributions in our paper:
\begin{itemize}
    \item We propose a novel OD and OC segmentation approach, which for the first time considers both the Cartesian domain and polar domain and explores the complementary.
    \item We design a Cartesian-polar dual-domain network (DDNet) with two encoding branches to learn rich contextual information from two domains for the joint OD and OC segmentation.
    \item We propose a dual-domain fusion module. It allows the element-wise fusion of the feature maps on different grids from two domains and enhances the expressive capability by learning the feature importance across two domains in element-wise.
\end{itemize}
\begin{figure*}
\begin{center}
\includegraphics[width=\textwidth]{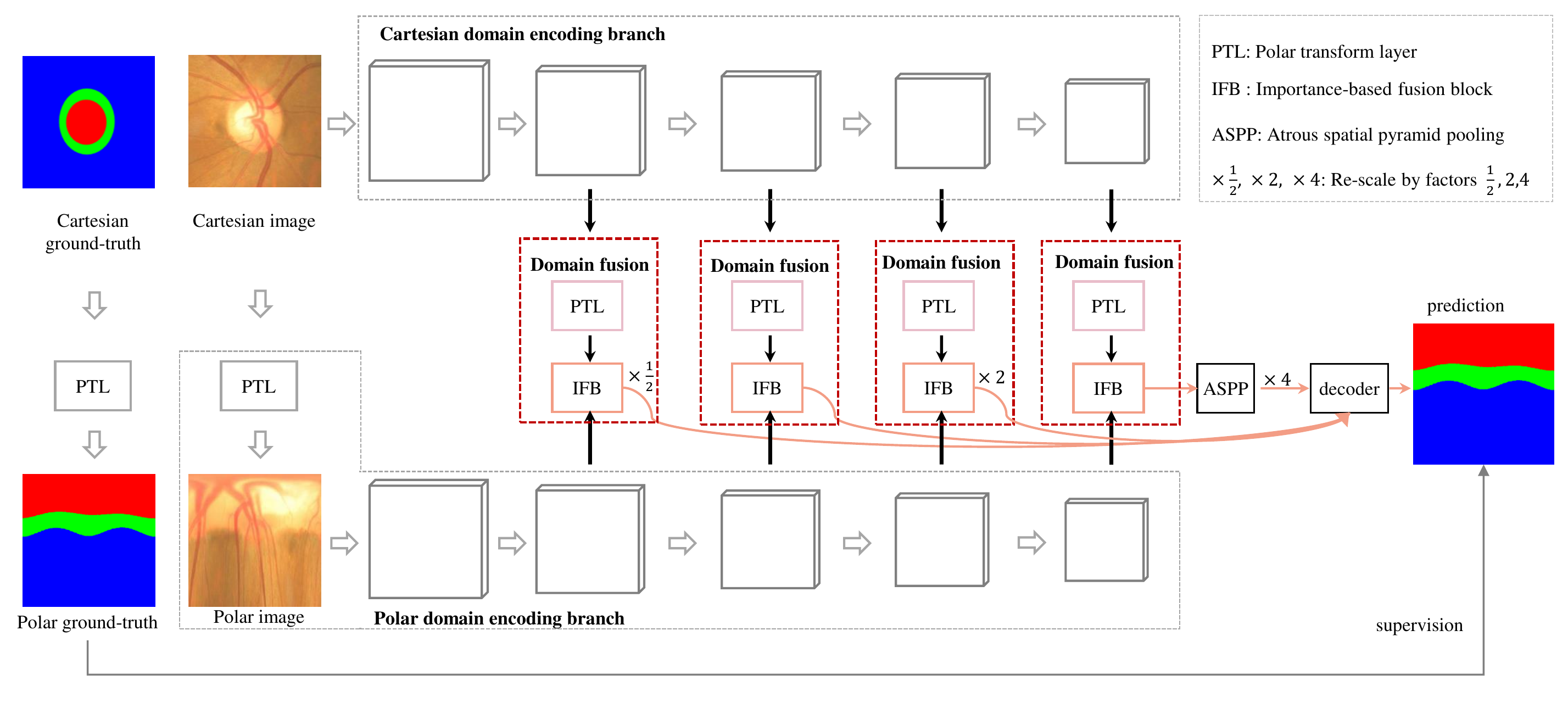}
   \caption{The architecture of the proposed DDNet. It involves four components. (1) The Cartesian domain encoding branch maps the image to feature maps on rectilinear grid. (2) The polar domain encoding branch maps the image to feature maps on polar grid. (3) The dual-domain fusion module builds the correspondence between two domains by the polar transform layer (PTL) and fuses the importance-refined features across domains in element-wise by the proposed importance-based fusion block (IFB). (4) The decoder aggregates the fused features from low-level to high-level and makes dense predictions. The atrous spatial pyramid pooling \cite{chen2018encoder} performed on the fused feature maps at the last stage is used to produce more scales of feature maps.}
\label{fig:framework}
\end{center}
\end{figure*}
\section{Related Works}
\label{sec:RelatedWork}
For the wide clinical applicability, the joint OD and OC segmentation has attracted much attention in the past decades. The OD and OC segmentation approaches update from hand-crafted to deep learning based while the domain performing segmentation extends from the Cartesian domain to the polar domain.
\\ \indent \textbf{Hand-crafted features based in Cartesian domain.} Most approaches are developed in the Cartesian domain with hand-crafted features. There are mainly two veins. One is based on the shape priors and tries to delineate the boundaries, \eg \cite{Lowell_TMI_2004, Aquino_TMI_2010_Hough, Yu_TITB_2012, Zilly_CMIG_2017, Li_Neurocomputing_2018, Bekkers_TPAMI_2018}. The other is based on the appearance priors and aims to distinguish the OD and OC pixels/regions from the background, \eg \cite{Cheng_TMI_2013, zheng2013optic, Xu_MICCAI_2014, Salazar_JBHI_2014}. Due to the limited expressive capability of the hand-crafted features, those methods are intractable to segment the subtle OC. They are also fragile when the OD is surrounded by bright exudates, peripapillary atrophy \etc.
\\ \indent \textbf{Deep features based in Cartesian domain.} More effective OD and/or OC segmentation approaches root in powerful representation learning methods within Cartesian domain. Deep Retinal Image Understanding (DRIU) \cite{Maninis_MICCAI_2016} takes the five-stage VGG16 \cite{VGG16} as the base network and learns feature maps for the OD segmentation. In \cite{Sevastopolsky_PRIA_2017} and \cite{DenseNet_Systems_2018}, variants of UNet \cite{UNet_MICCAI_2015} were proposed to segment the OD and OC. Nevertheless, the deep learning based segmentation approaches in Cartesian domain encounter difficulties in learning due to the imbalanced class distributions and the small classes such as OC pixels are prone to be misclassified.
\\ \indent \textbf{Deep features based in polar domain.} Most recently, MNet \cite{Fu_TMI_2018}, performing segmentation in polar domain, was proposed. It learns representations directly from polar images by a simplified UNet \cite{Fu_TMI_2018} with multiscale inputs and produces polar segmentation maps. Then an inverse polar transform is used to map the polar segmentation results back to Cartesian domain. Although MNet \cite{Fu_TMI_2018} improves the segmentation performances significantly, it ignores the contextual information from Cartesian domain. 
\\ \indent Different from the previous approaches, our DDNet learns representations from dual domains for the joint segmentation of the OD and OC. To the best knowledge of the authors, this is the first segmentation network designed to explore the complementary of the Cartesian domain and polar domain.
\section{Cartesian-polar Dual-domain Network}
\label{sec:ProposedMethod}
\indent The proposed Cartesian-polar dual-domain segmentation network (DDNet) roots in the two-branch of domain feature encoder and well-designed dual-domain fusion module. With the rich representations learned from two domains, the DDNet makes dense predictions by a decoder.
\subsection{Network Architecture}
The Cartesian domain is the original domain that both natural images and fundus images are captured. In this domain, the rectilinear grid is used and the geometry structures are well visualised. Naturally, the end-to-end segmentation networks \cite{long2015fully,chen2014semantic,Noh_deconv_seg_2015,SegNet_2017,chen2018encoder} were developed for natural images in Cartesian domain. By directly transferring the great successes achieved in natural image segmentation to fundus image segmentation, U-shaped networks \cite{Maninis_MICCAI_2016,Sevastopolsky_PRIA_2017,DenseNet_Systems_2018} were proposed. To alleviate the imbalanced class distributions among the OC, rim, and background pixels, MNet \cite{Fu_TMI_2018} was proposed to segment the OC and OD in the polar domain. However, the contextual information learned from single domain is still insufficient. Therefore, it is highly desired to learn richer representations for the segmentation task.
\\ \indent We observe that the shapes and spatial layouts of the OC, rim, OD, and vessels are completely different in Cartesian domain and polar domain. For example, the OC is ellipse-like and the rim is ringlike in Cartesian domain while they are band-like in polar domain. The vessels in Cartesian domain extend radially from superior and inferior to the OD while they are almost vertical layout in polar domain. Also by transforming the image in Cartesian domain to polar domain, the structures close to the transformation origin in Cartesian domain are amplified in polar domain while the structures far away from the transformation origin are squeezed. Such evident differences imply that complementary contextual information is embedded in two domains. Learning representations from both domains and fusing them results in richer representations. To this end, we propose the DDNet. 
\\ \indent Fig. \ref{fig:framework} illustrates the architecture of our DDNet. It involves the following four components:
\begin{itemize}
    \item \textbf{Cartesian domain encoding branch.} This branch maps the Cartesian image $X$ into the feature maps on rectilinear grid. We directly use the modified Xception model \cite{chen2018encoder} to learn the feature representations since it has shown promising performances in natural scene semantic segmentation. It can be divided into five stages according to the spatial sizes of the feature maps. For convenience, we denote the feature maps by the last convolutional layer at the $l$-th stage as $f^{(l)}(X)$.
    \item \textbf{Polar domain encoding branch.} This branch first maps the Cartesian image $X$ on rectilinear grid to polar image $X_{polar}$ on polar grid by a polar transform layer (PTL), then forwards the polar image into the modified Xception model \cite{chen2018encoder} and generates feature maps on polar grid. For convenience, we denote the feature maps by the last convolutional layer at the $l$-th stage as $g^{(l)}(X)$.
    \item \textbf{Dual-domain fusion module.} This module is designed to incorporate $f^{(l)}(X)$ on rectilinear grid and $g^{(l)}(X)$ on polar grid. It first transforms the feature maps $f^{(l)}(X)$ on the rectilinear grid to feature maps  $f_{polar}^{(l)}(X)$ on the polar grid by a PTL. Then it fuses the feature maps $f_{polar}^{(l)}(X)$ and $g^{(l)}(X)$ on the polar grid in element-wise by the important-based fusion block (IFB). Its detailed description will be given in next subsection.
    \item \textbf{Decoder.} The decoder first re-scales the fused feature maps to a unified spatial size, then concatenates them and makes dense predictions to obtain the segmentation maps on polar grid.
\end{itemize}
\indent The segmentation maps by our DDNet are on polar grid. The joint segmentation problem is formulated as a three classes (i.e., OC, rim, background) dense classification and the cross-entropy loss is used. In the training phase, our DDNet is supervised by the segmentation ground-truth in polar domain, which is transformed from the Cartesian ground-truth by the PTL. During the testing phase, an extra inverse polar transformation is performed on the outputs of the DDNet to obtain the final segmentation results on rectilinear grid.
\begin{figure*}
\includegraphics[width=\textwidth]{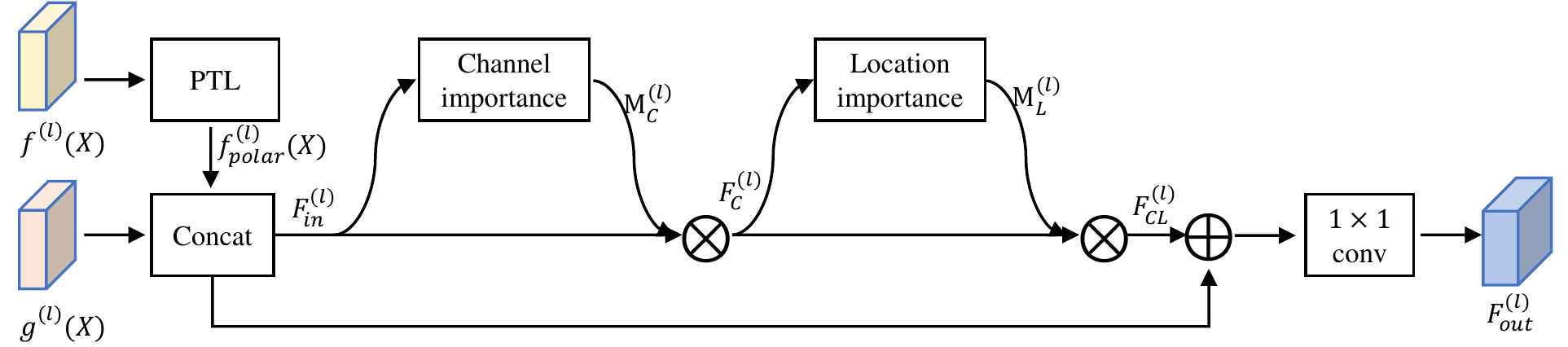}
   \caption{The dual-domain fusion module. As illustrated, the PTL builds the correspondence between rectilinear grid and polar grid. It transforms the feature maps $f^{(l)}(X)$ on rectilinear grid from Cartesian domain encoding branch to $f^{(l)}_{polar}(X)$ on polar grid. The channel importance map $M^{(l)}_{C}$ and location importance map $M^{(l)}_{L}$ together encode the importance of the feature in element-wise. Finally, a $1\times 1$ convolutional layer fuses the features across the channel dimension and generates the fused feature maps $F^{(l)}_{out}$.}
\label{fig:AttFusion}
\end{figure*}
\subsection{Dual-domain Fusion Module}
Given the feature maps on two different grids, the dual-domain feature fusion for dense prediction pursuits a weighted strategy in element-wise that best fits the segmentation ground-truth. It is required to build the correspondence between two domains and exploit the complementary information for the element-wise classification. Fig. \ref{fig:AttFusion} illustrates the dual-domain fusion module.
\\ \indent The Cartesian domain branch takes the image on the rectilinear grid as input and outputs feature maps on rectilinear grid. Differently, the polar domain branch first transforms the image on the rectilinear grid to that on the polar grid, then outputs the feature maps on the polar grid. To build the correspondence of the feature maps on different grids, we adopt the PTL \cite{PTN_ICLR} and transform the feature maps on the rectilinear grid to that on polar grid. Formally, denoting the feature maps with $K^{(l)}$ channels and size of $H^{(l)} \times W^{(l)}$ in Cartesian domain from the $l$-th stage as $f^{(l)}(X)$ and the point coordinates on rectilinear grid as $(u,v)$, the PTL \cite{PTN_ICLR} adopts the differentiable image sampling technique \cite{jaderberg2015spatial} and outputs the sampled polar feature maps $f_{polar}^{(l)}(X)$ with the same spatial size and channel whose point coordinates on polar grid are denoted as $(\xi, \theta)$. In terms of $(u,v)$ and $(\xi,\theta)$, the PTL is expressed as:
\begin{align}
\label{eq:PTL}
\begin{cases}
\xi = 2\sqrt{((u-u_0)^2 + (v-v_0)^2)},\\
\theta = \atantwo (u-u_0, v-v_0),
\end{cases}
\end{align}
where $(u_o, v_o)$ is the centre point of $f^{(l)}(X)$ and $u_o = H^{(l)}/2, v_o=W^{(l)}/2$.
\\ \indent To select the informative features from two domains for each element and enhance the expressive capability of the representations, the importance of the feature across the domains is learned. It is implemented by learning two important matrices $M^{(l)}_C$ and $M^{(l)}_L$. $M^{(l)}_C$ is the channel importance map, encoding the importance of the feature map from two domains. $M^{(l)}_L$ is the location importance map, encoding the importance of the spatial location. 
\\ \indent Formally, taking the feature maps $f_{polar}^{(l)}(X)$ and $g^{(l)}(X)$ from the $l$-stage of the Cartesian domain encoding branch and polar domain encoding branch respectively as input $F_{in}^{(l)} = [f_{polar}^{(l)}(X), g^{(l)}(X)]$, the importance weighted feature maps are obtained by:
\begin{align}
F_{C}^{(l)} & = M_{C}(F_{in}^{(l)}) \otimes F_{in}^{(l)}, \\ \nonumber
F_{CL}^{(l)} & = M_{L}(F_{C}^{(l)}) \otimes F_{C}^{(l)},
\end{align}
where $\otimes$ is the element-wise multiplication, $F_{C}^{(l)}$ is the feature maps weighted by $M^{(l)}_C$, and  $F_{CL}^{(l)}$ is the final output. Partially inspired by the CBAM \cite{woo2018cbam} which is designed to learn a channel attention map and spatial attention map at each CNN stage, we adopt the same implementation to learn $M^{(l)}_C$ and $M^{(l)}_L$. Finally we add the important weighted feature maps $F_{CL}^{(l)}$ and $F_{in}^{(l)}$, and use a $1\times 1$ convolutional layer to fuse the feature maps across the channel dimension:
\begin{align}
F_{out}^{(l)} & = (F_{CL}^{(l)} + F_{in}^{(l)}) \ast \mathbf{w}_{fusion}^{(l)},
\end{align}
where $\mathbf{w}_{fusion}^{(l)}$ is convolutional weights of the $1\times 1$ convolutional layer.
\subsection{Analysis}
From the view of representation theory, our proposed DDNet not only learns representations with powerful discriminativeness, but also benefits from the translation equivariant and rotation equivariant that the two-branch of domain encoder achieves respectively.
\\ \indent Essentially, the Cartesian domain branch learns feature representations by performing translational convolutions on a translation symmetry group \cite{cohen2016group} \cite{kondor2018generalization}. Consequently, the feature representations $f^{(l)}(X)$ for the input image $X$ achieve to translation equivariance and satisfy:
\begin{align}
\label{eq:translation}
    f^{(l)}(L_t \circ X) = L_t \circ f^{(l)}(X),
\end{align}
where $L_t$ is a translation action. This means that performing a translation $L_t$ on the input image $X$ (forming $L_t \circ X$) and then passing it through the Cartesian domain encoding branch will give the same result as first forwarding the input image $X$ to the Cartesian domain encoding branch (forming $f^{(l)}(X)$) and then performing the same translation $L_t$ on the learned representation. More specifically, as is exampled in Fig. \ref{fig:exam4two_sys}(b), for any pixel $q_1\in X$, denoting the corresponding pixel in the translated image $L_t \circ X$ as $L_t \circ q_1$, the Cartesian domain encoding branch maps them to a same representation, i.e., $f^{(l)}(q_1) = f^{(l)}(L_t \circ q_1)$. In other words, the representations for each pixel by the Cartesian domain encoding branch are invariant to translation and beneficial to the pixel-wise classification.
\\\indent Different from the Cartesian domain encoding branch, the polar domain encoding branch learns feature representations by performing rotation convolutions on a rotation symmetry group $SO(2)$ \cite{PTN_ICLR}. Formally, we model the feature representations $g^{(l)}(X)$ from the polar domain encoding branch as two mappings $g^{(l)}(X)=h^{(l)}(PTL(X))$. The first one is the PTL, which maps the Cartesian image to polar and reduces the rotation action on the Cartesian image to the translation action on the corresponding polar image. This implies that, performing a rotation action $L_r$ on $X$, there exists a translation $L_t$ satisfying:
\begin{align}
\label{eq:rotation2translation}
    PTL(L_r \circ X) = L_t\circ PTL(X).
\end{align}
The second one corresponds to the convolutional neural network, which maps the polar image $PTL(X)$ to feature representation $g^{(l)}(X)$. With Eq. \ref{eq:rotation2translation}  and Eq. \ref{eq:translation}, the feature representations $g^{(l)}(X)$ from the polar domain encoding branch satisfy:
\begin{align}
\label{eq:translation}
    g^{(l)}( L_r \circ X ) & = h^{(l)}(PTL(L_r \circ X) )  \\ \nonumber
    & = h^{(l)}(L_t\circ PTL(X)) \\ \nonumber
    & = L_t \circ h^{(l)}(PTL(X)) \\ \nonumber
    & = L_t \circ g^{(l)}(X).
\end{align}
This means that performing a rotation on the input image $X$ (forming $L_r \circ X$) and then forwarding it to the polar domain encoding branch will give the same result as first forwarding the input image $X$ to the polar domain encoding branch (forming $g^{(l)}(X)$) and then performing a translation $L_t$ on the learned feature representation. More specifically, as is exampled in Fig. \ref{fig:exam4two_sys}(c), for any pixel $q_2$ in $X$ and the corresponding pixel $L_r\circ q_2$ in the rotated image $L_r\circ X$, the polar domain encoding branch maps them to a same feature vector, i.e., $g^{(l)}( q_2 )=g^{(l)}( L_r \circ q_2 )$. In other words, the representations for each pixel by the polar domain branch are invariant to rotation and beneficial to the pixel-wise classification.
\\ \indent By fusing the translation invariant representations and rotation invariant representations, our DDNet obtains more powerful representations. Next, we will demonstrate its effectiveness by experiments. 
\section{Experimental Results}
\label{sec:exp}
The segmentation performances of our DDNet are first evaluated and compared on the public dataset ORIGA \cite{ORIGA} for the OD and OC segmentation. Then we apply it to the CDR estimation. The ORIGA \cite{ORIGA} contains 650 fundus images size of $3072\times2048$. 325 images are used as training images including 73 glaucoma cases and 325 images are used as testing images including 95 glaucoma cases. For each image in ORIGA \cite{ORIGA}, the segmentation masks of the OD and OC, the CDR value by experts are provided. 
\subsection{Implementation Details}
\indent \textbf{Data Augmentation.} In the training phase, the images from the training set are flipped horizontally in a random way and scaled by a random factor ranging from 0.9 to 1.1. The OD region only takes a small region in the retinal fundus image. To crop a small sized OD window, we simply train an OD segmentor by finetuning the pre-trained Deeplabv3+ \cite{chen2018encoder}. According to the OD mask from the OD segmentor, we calculate the OD centre and crop an $800\times800$ window as the inputs of the proposed DDNet.
\\ \indent \textbf{Training.} The proposed DDNet is built on the Deeplabv3+ \cite{chen2018encoder}. A two-stage training procedure is adopted. First, we pre-train the parameters in the Cartesian domain encoding branch and polar domain encoding branch separately by finetuning the pre-trained Deeplabv3+ \cite{chen2018encoder} on Pascal VOC and obtain the single domain segmentation model, denoted as Deeplabv3+ (Cartesian) and Deeplabv3+ (polar)  respectively. Then, the parameters in the whole DDNet are trained. The hyper-parameters include: the mini-batch size (4), learning rate (0.007 in the first stage, 0.001 in the second stage), maximum number of training iterations (10,000), momentum (0.9) and weight decay (0.00004). 
\\ \indent \textbf{Testing.} At the testing phase, the whole input image is first forwarded to the OD segmentor, and the OD window size of $800\times800$ is cropped. Then we forward the OD window to the trained DDNet and obtain the segmentation masks of OD and OC on polar grid. Finally, with an inverse polar transform, we obtain the final segmentation masks on rectilinear grid.
\begin{table*}[h]
	\setlength{\belowcaptionskip}{-15pt}
	\begin{center}
		\begin{tabular}{c| c | c c c c c}
			\hline \multicolumn{2}{c|}{Methods} & ~~$E_{disc}$~~ & ~~$BLE_{disc}/std$~~ & ~~$E_{cup}$~~ & ~~$BLE_{cup}/std$~~ & ~~$E_{rim}$~~ \\ \hline
			\multicolumn{1}{c|}{\multirow{4}{*}{hand-crafted}}	&	R-bend \cite{Joshi_TMI_2011}  &  0.129  &  -  &  0.395  &  -  &  - \\ 
			& ASM \cite{ASM}  &  0.148  &  -  &  0.313  &  -  &  -  \\
			& Superpixel \cite{Cheng_TMI_2013}  &  0.102  &  -  &  0.264  &  -  &  0.299  \\
			& LRR \cite{Xu_MICCAI_2014}  &  -  &  -  &  0.244  &  -  &  -  \\\hline
			\multicolumn{1}{c|}{\multirow{6}{*}{{deep~learning}}} & lightweight U-Net \cite{Sevastopolsky_PRIA_2017}  &  0.115  &  -  &  0.287  &  -  &  0.303 \\
			&   FC-DenseNet \cite{DenseNet_Systems_2018}  &  0.067  &  -  &  0.231  &  -  &  -  \\
			&   MNet \cite{Fu_TMI_2018}  &  0.071  &  6.70/6.93  &  0.230  &  14.38/9.96  &  0.233  \\
			&   DeepLabv3+ \cite{chen2018encoder} (Cartesian)  &  0.059  &  5.51/3.79
			&  \textcolor{blue}{0.209}  &  \textcolor{blue}{12.93/8.39}  &  0.212  \\
			&   DeepLabv3+ \cite{chen2018encoder} (polar)  &  \textcolor{blue}{0.057}  &  \textcolor{blue}{5.26/3.38}
			&  0.214  &  13.23/9.09  &  \textcolor{blue}{0.210}  \\
			&   \textbf{DDNet (ours)}  &  \textcolor{red}{0.054}  & \textcolor{red}{5.01/3.35}  &  \textcolor{red}{0.204}  &  \textcolor{red}{12.48/8.39}  &  \textcolor{red}{0.201} \\\hline
		\end{tabular}
		\caption{Performance comparisons of the different methods on ORIGA \cite{ORIGA}. $E$ denotes the overlapping error. $BLE/std$ are the boundary location error and its standard deviation. The best results and second best results are marked in red and blue respectively. (Best viewed in colour) }\label{tab:PerformanceComparisonORIGA}
	\end{center}
\end{table*}
\begin{figure*}[h]
\setlength{\belowcaptionskip}{-15pt}
\includegraphics[width=\textwidth]{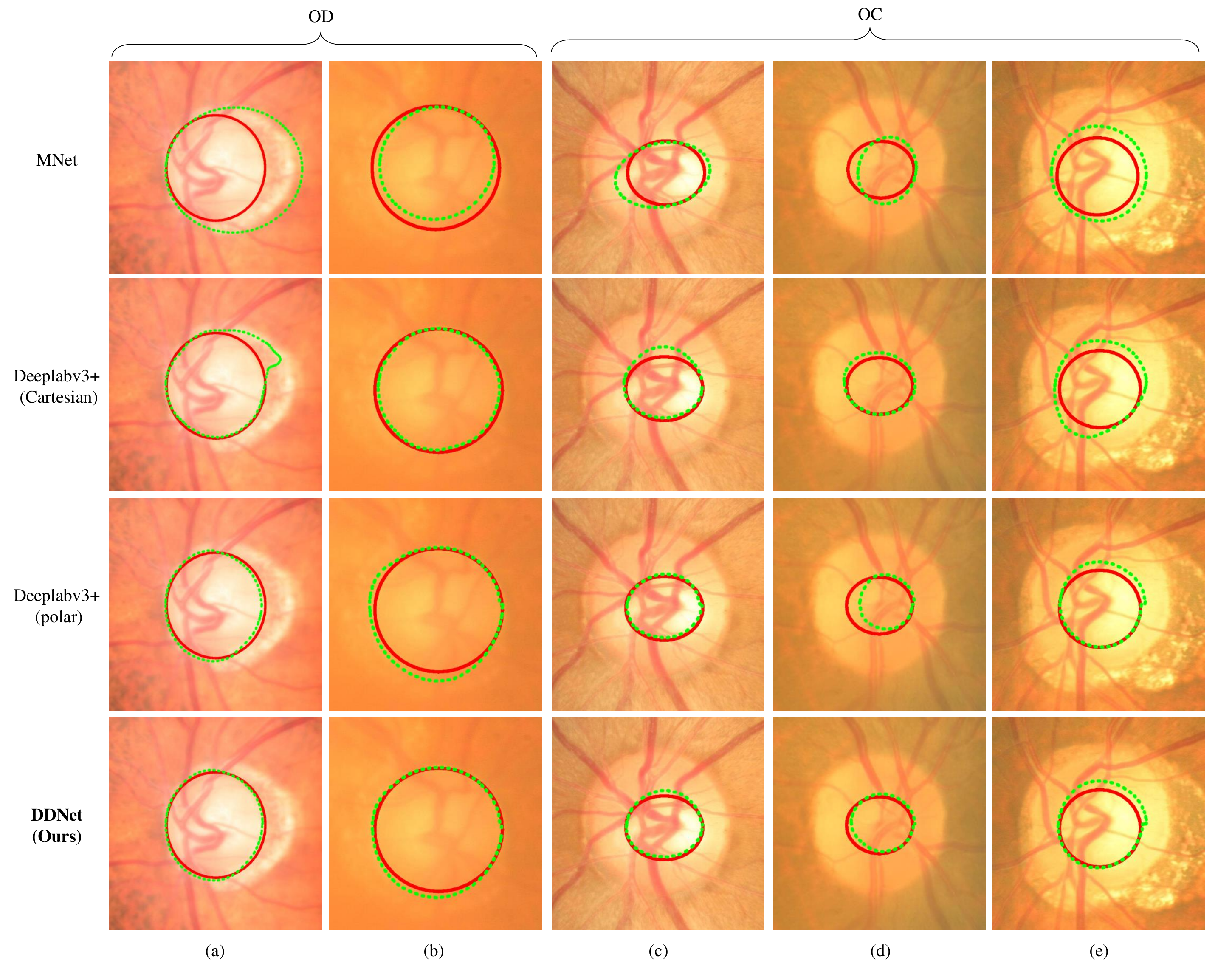}
\caption{OD and OC segmentation results on ORIGA \cite{ORIGA} dataset: (a) and (b) are challenging images for OD segmentation and (c) $\sim$ (e) are challenging images for OC segmentation. From top to bottom: results by MNet \cite{Fu_TMI_2018}, Deeplabv3+ \cite{chen2018encoder} trained in Cartesian domain, Deeplabv3+ \cite{chen2018encoder} trained in polar domain and our DDNet. The solid contours in red and the dashed green contours are delineated by experts and the segmentation approaches respectively. (Best viewed in colour) }
\label{fig:examples}
\end{figure*}
\begin{figure}[h]
\centering
\includegraphics[width=0.62\linewidth]{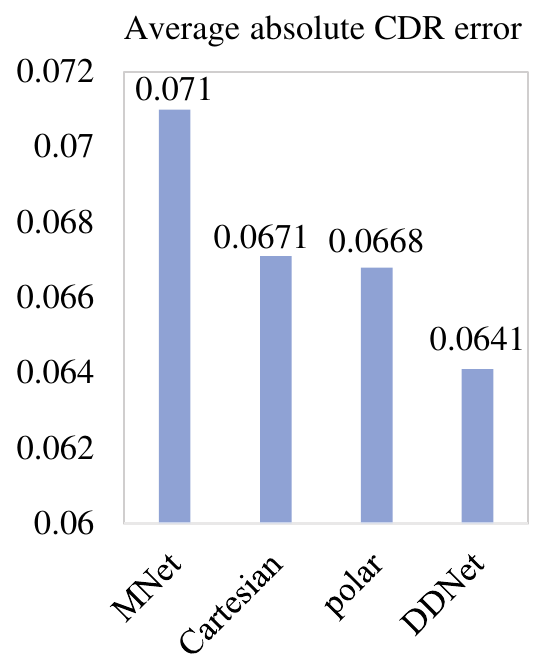}
\caption{Performance comparisons on the CDR estimation. From left to right: the average absolute CDR errors by MNet \cite{Fu_TMI_2018}, Deeplabv3+ \cite{chen2018encoder} (Cartesian), Deeplabv3+ \cite{chen2018encoder} (polar) and our proposed DDNet.}
\label{ScrreningORIGA}
\end{figure}
\subsection{Segmentation Performances}
\indent We adopt the overlapping error $E$ and the average boundary location error $BLE$ to evaluate the performances as introduced in \cite{Fu_TMI_2018} and \cite{Chakravarty_CMPB_2017} respectively. The former measures the ratio of number of pixels that are wrongly classified to the number of pixels in the union region of the segmentation mask and the ground-truth. The later measures the average absolute distance between the boundaries $\Hat{B}$ of the segmentation mask and the boundaries $B$ of the ground-truth:
\begin{align}
    BLE(\Hat{B}, B)=\frac{1}{n}\sum_{\alpha\in \{ \alpha_1,\cdots,\alpha_n \}} | d(\Hat{B}, \alpha) - d(B, \alpha) | \; ,
\end{align}
where $d(\cdot, \alpha)$ is the Euclidean distance of the boundary point in the direction $\alpha$ to the centroid of the target, and $\{ \alpha_1,\cdots,\alpha_n \}$ is the set of the uniformly sampled directions. As same as the setting used in \cite{Chakravarty_CMPB_2017}, $n$ is set to $24$.
\\ \indent We compare the segmentation performances of the proposed DDNet with R-bend \cite{Joshi_TMI_2011}, ASM \cite{ASM}, Superpixel \cite{Cheng_TMI_2013}, LRR \cite{Xu_MICCAI_2014}, lightweight U-Net \cite{Sevastopolsky_PRIA_2017}, FC-DenseNet \cite{DenseNet_Systems_2018}, DeepLabv3+ \cite{chen2018encoder} (Cartesian) and DeepLabv3+ \cite{chen2018encoder} (polar). Among them, the first four are hand-crafted features based. The last five are deep features based. 
\\ \indent We report the performance comparisons in Table. \ref{tab:PerformanceComparisonORIGA}, in which the overlapping errors of the OD, OC, and rim are denoted as $E_{disc}$, $E_{cup}$ and $E_{rim}$, respectively. The average boundary location errors of OD and OC are denoted as $BLE_{disc}$ and $BLE_{cup}$, respectively. It is observed that: (1) the proposed DDNet achieves the lowest overlapping errors as well as the boundary location errors; (2) Compared to the MNet \cite{Fu_TMI_2018} which is specifically designed for the joint OD and OC segmentation, our DDNet outperforms it by $1.7\%$, $2.6\%$ and $3.2\%$ with the overlapping errors of the OD, OC and rim respectively and by $1.44$ and $1.9$ with the boundary location errors of the OD and OC respectively; (3) Compared to the Deeplabv3+ \cite{chen2018encoder} performed in Cartesian domain, our DDNet reduces the overlapping errors of the OD, OC and rim by $0.5\%$, $0.5\%$ and $1.1\%$ respectively; (4) Compared to the Deeplabv3+ \cite{chen2018encoder} performed in polar domain, our DDNet reduces the overlapping errors of the OD, OC and rim by $0.3\%$, $1.0\%$ and $0.9\%$ respectively.
\\ \indent The segmentation results by MNet \cite{Fu_TMI_2018}, DeepLabv3+ \cite{chen2018encoder} (Cartesian), DeepLabv3+ \cite{chen2018encoder} (polar) and the proposed DDNet are illustrated in Fig. \ref{fig:examples}. It is illustrated that the Deeplabv3+ \cite{chen2018encoder} (polar) achieves superior results in Fig. \ref{fig:examples}(a) and Fig. \ref{fig:examples}(c) but inferior results in Fig. \ref{fig:examples}(b) and Fig. \ref{fig:examples}(d) to Deeplabv3+ \cite{chen2018encoder} (Cartesian). By fusing the features extracted from two domains, our DDNet is able to achieve the superior results. Compared to MNet \cite{Fu_TMI_2018}, our DDNet achieves more accurate segmentation results. The last column shows a challenging example that all methods fail to segment the OC. 
\subsection{Application on the CDR Estimation}
Glaucoma is the first leading cause of irreversible vision impairment and blindness \cite{tham2014global}. In clinical, the diagnosis of glaucoma commonly relies on multiple measures such as the CDR, the visual field and intraocular pressure, \etc. Generally, the larger the CDR is, the higher risk the patient is at. In what follows, we estimate the CDR according to the segmentation masks of OD and OC. 
\\ \indent The CDR value is defined as the ratio of the vertical diameter of the OC to the vertical diameter of the OD. To evaluate the performances on the CDR estimation of the segmentation approaches, we follow \cite{Fu_TMI_2018} and adopt the absolute error. Fig. \ref{ScrreningORIGA} shows the results of MNnet \cite{Fu_TMI_2018}, Deeplabv3+ \cite{chen2018encoder} (Cartesian), Deeplabv3+ \cite{chen2018encoder} (polar), and our DDNet. Obviously, our DDNet gets the lowest absolute CDR estimation error $0.0641$. The Deeplabv3+ \cite{chen2018encoder} achieves similar absolute CDR errors in Cartesian domain ($0.0671$) and polar domain ($0.0668$), and both of them are superior to the MNet ($0.071$) \cite{Fu_TMI_2018}.
\section{Conclusion}
\label{sec:Conclusion}
This paper focuses on the joint segmentation of OD and OC in retinal fundus images. Due to the absence of depth, representations learned from single domain are insufficient to partition of the OC and rim. To improve performances, we propose to learn representations from both the Cartesian domain and polar domain, and present the Cartesian-polar Dual-domain segmentation network (DDNet). On one hand, our DDNet benefits from the complementary contextual information exploited from images in Cartesian domain and polar domain. On the other hand, our DDNet benefits from the translation equivariance achieved by the CNNs in Cartesian domain and the rotation equivariance achieved by the CNNs in polar domain. By fusing the representations from both domains, the representations by our DDNet are more powerful. We also validate the-state-of-the-art segmentation performances of the DDNet on ORIGA \cite{ORIGA}. When applying the DDNet to the CDR estimation, it achieves lowest absolute error, which demonstrates the potential application on glaucoma screening. Our DDNet benefits from the complementary of two domains. But it is still an open question that how to fuse the feature maps such that the fused features are equivariant to translation and rotation. This will be our future work.
{\small
	\bibliographystyle{ieee}
	\bibliography{egbib1}
}
\end{document}